\documentclass[
]{ceurart}

\usepackage{listings}
\usepackage{url}
\usepackage[center]{caption}

\begin{document}

\copyrightyear{2022}
\copyrightclause{Copyright for this paper by its authors.
  Use permitted under Creative Commons License Attribution 4.0
  International (CC BY 4.0).}

\conference{Forum for Information Retrieval Evaluation, December 9-13, 2022, India}

\title{Implementing Deep Learning-Based Approaches for Article Summarization in Indian Languages}


\author[1,3]{Rahul Tangsali}
\cormark[1]
\fnmark[1]
\address[1]{SCTR's Pune Institute of Computer Technology, Pune}
\address[2]{Indian Institute of Technology Madras, Chennai}
\address[3]{L3Cube, Pune}

\author[1,3]{Aabha Pingle}
\fnmark[1]

\author[1,3]{Aditya Vyawahare}
\fnmark[1]

\author[1,3]{Isha Joshi}
\fnmark[1]

\author[2, 3]{Raviraj Joshi}


\begin{abstract}
The research on text summarization for low-resource Indian languages has been limited due to the availability of relevant datasets. This paper presents a summary of various deep-learning approaches used for the ILSUM 2022 Indic language summarization datasets. The ISUM 2022 dataset consists of news articles written in Indian English, Hindi, and Gujarati respectively, and their ground-truth summarizations. In our work, we explore different pre-trained seq2seq models and fine-tune those with the ILSUM 2022 datasets. In our case, the fine-tuned SoTA PEGASUS model worked the best for English, the fine-tuned IndicBART model with augmented data for Hindi, and again fine-tuned PEGASUS model along with a translation mapping-based approach for Gujarati. Our scores on the obtained inferences were evaluated using ROUGE-1, ROUGE-2, and ROUGE-4 as the evaluation metrics.
\end{abstract}

 
\begin{keywords}
  Abstractive text summarization \sep
  Indian Languages \sep
  NLP \sep
  Pretrained models
\end{keywords}

\maketitle

\section{Introduction}

Text summarization is a trending research domain that has gained popularity with a plethora of emerging use cases seeking its application \cite{Vhatkar2020SurveyOT, article_dehru}. The last few decades have witnessed tremendous growth in NLP research, especially text summarization. Text summarization has applications in a wide range of domains, including medicine, politics, news, etc. With the massive influx of news data in the form of newspaper articles, digital media, social media platforms, and so on, a need exists to automate the news summarization process so that useful insights could be achieved much faster than human workers were employed for the same task. Effective summarization approaches investigated recently have hastened the process and made their mark in the NLP research community by achieving state-of-the-art (SoTA) accuracies.

\indent Three distinct types of text summarization techniques like extractive, abstractive, and hybrid. In extractive text summarization, key sentences and phrases are picked from the original document and are integrated to generate a final summary \cite{inproceedings_extractive_survey}. This summarization technique is easier to perform, but it may overlook the text's overall context or omit some essential information. This type of summary text is helpful for taking notes. Abstractive summarization analyses the full text and generates a summary based on the fundamental concepts of the text \cite{inproceedings_abstractive_survey}. This summary is made using an entirely different wording style than the original text. Unlike the extractive summarization methodology, sentences from the original text aren't picked up directly. Abstractive summarization provides an intelligently curated summarization using unique phrases which are not native to the input text. However, with deep learning methodologies, preparing abstractive summaries could be difficult and take a long time with human judgment. The hybrid-based text summarization approach utilizes both extractive and abstractive text summarization methods to generate the final summary \cite{inbook_hybrid, SAHOO20181228}.

\indent With the emergence of NLP research worldwide, research on text summarization has been conducted in high-resource languages such as English and texts written in Indian subcontinent-based languages. Hindi and Gujarati are two of the most spoken Indian languages. Hindi is the most spoken language in India and is considered the official language in 9 states and 3 union territories and an additional official language in 3 other states across the country. Hindi is also one of the 22 scheduled languages of the Republic of India. Hindi is spoken by approximately 615 million people worldwide and was recorded as the third most spoken language in the world as of 2019. Gujarati is an Indo-Aryan language spoken predominantly by the Gujarati people in the Indian state of Gujarat. It is the sixth most spoken language in India and is spoken by around 55 million people worldwide. Hindi and Gujarati are spoken by a considerable percentage of the population across the world. Yet, there has been a backfoot witnessed in NLP research in these languages compared to high-resource languages spoken worldwide.

\indent Text summarization research stretches back to 1958 when the first paper on the subject was published \cite{luhn_summarization}. Since then, various methodologies have been presented for both abstractive and extractive text summarization in English. These include statistical-based, clustering-based, graph-based, semantic-based, machine learning, and deep learning-based approaches. Deep learning-based approaches, which focus on training neural nets, include work done by Mohsen et al. \cite{mohsen2020hierarchical}, Xu \cite{xu2019neural}, Alami et al. \cite{alami2019enhancing}, and Anand and Wagh \cite{anand2019effective}. In addition, encoder-decoder models have been proposed, with attention mechanisms incorporated in several proposed methodologies. 

\indent In comparison to English, lesser research has been done on text summarization research in Hindi and Gujarati. There is a significant shortage in dataset resources, preprocessing methodologies, and other research for many Indian languages, especially Gujarati, compared to English. This motivated us to develop system pipelines that could perform efficient extractive summarization for articles written in Hindi and Gujarati and achieve decent accuracy for the generated summaries. Many organizations are leveraging their services to Indian language speakers, and we aim to solve a small part of this challenge by performing summarization research in two of the widely spoken languages in India.

\indent We implement pre-trained models \cite{HAN2021225} and tweak the conventional pipelines along with fine-tuning with new data to obtain better results than previously implemented systems. For English, we implement the PEGASUS\footnote{\url{https://huggingface.co/l3cube-pune/english-pegasus-summary}} \cite{10.5555/3524938.3525989}, BRIO \cite{liu-etal-2022-brio}, and T5 \cite{raffel2020exploring} models and also leverage the SentenceBERT model for extractive summarization purposes \cite{reimers2019sentence, liu2019fine}. For Hindi, we implemented fine-tuning of IndicBART\footnote{\url{https://huggingface.co/l3cube-pune/hindi-bart-summary}} \cite{dabre-etal-2022-indicbart} with a right-shift operation (augmenting the original dataset by shifting the last sentence of the article to the top), XL-Sum \cite{hasan-etal-2021-xl}, and mBART \cite{liu2020multilingual} models. For Gujarati, we implemented extractive summarization by translating each sentence in the Gujarati article to English, and by creating a corresponding mapping between the Gujarati and translated English sentences, and applying fine-tuned PEGASUS model for English to the resultant English article to generate the English summary. The generated extractive summary in English is then translated back to Gujarati by a back-mapping mechanism to get the final Gujarati summary. We also fine-tuned XL-Sum and mBART\footnote{\url{https://huggingface.co/l3cube-pune/gujarati-bart-summary}} models for Gujarati article summarization.

\section{Related Work}

Text summarization research dates back to 1958 when the first article on the topic \cite{luhn_summarization} was published. Since then, numerous rule-based and deep learning-based techniques have been presented. Rule-based approaches include work done by Baxendale \cite{5392648}, which selects sentences for a summary based on word position and heading of the article, and that by Oliviera in 2016 \cite{10.1016/j.eswa.2016.08.030}, which used scoring criteria such as lexical similarity, sentence centrality, text rank, and so on for text summarization.
\\ \indent Research on deep-learning approaches for text summarization picked up the pace when encoder-decoder \cite{https://doi.org/10.48550/arxiv.2110.15253} and attention-based architectures \cite{NIPS2017_3f5ee243} were proposed. Yu \cite{Yu2017SummarizationWA} suggested methods for creating one-sentence summaries of news stories that use recurrent neural network models like LSTM \cite{LSTM_1} and GRU \cite{https://doi.org/10.48550/arxiv.1412.3555}, as well as with/without attention. In recent years, fine-tuning pre-trained models using domain-specific datasets has been the dominant paradigm in text summarization research. Pre-trained models which implement the BART \cite{lewis-etal-2020-bart}, T5 \cite{raffel2020exploring}, etc. architectures have been proposed, which are available in the Hugging Face library. Recent research includes the implementation of an importance-based ordering approach implemented by Zhao et al., a cascade approach to abstractive summarization with content selection and fusion proposed by Lebanoff et al \cite{lebanoff-etal-2020-cascade}., and usage of prompt-based models such as GPT-3 \cite{NEURIPS2020_1457c0d6}, PaLM \cite{https://doi.org/10.48550/arxiv.2204.02311}, T0 \cite{sanh2022multitask}, etc. Many times, articles considered for summarization can be multi-document in nature. Wang et al. \cite{wang-etal-2020-heterogeneous} suggested a task-specific architecture for multi-document summarization by combining numerous texts into a single graph. Zhong et al. \cite{zhong-etal-2020-extractive} implemented a semantic-based framework for the same.
\\ \indent In the case of Hindi and Gujarati, there has been relatively little research on text summarization. K. Vimal Kumar et al. \cite{10.1007/978-81-322-2250-7_29} suggested a graph-based method for summarising text in Hindi. Gulati et al. \cite{Gulati2017ANT} developed a unique fuzzy inference method for summarising multi-source Hindi literature. Gupta et al. \cite{Gupta2016TextSO} suggested a rule-based method for Hindi that included dead phrase and deadwood reduction strategies. Jain et al. \cite{article_jain} presented a real coded genetic algorithm for Hindi text summarization. For Gujarati, Shah and Patel suggested Gujarati Text Summarizer, which uses Textblob\footnote{https://textblob.readthedocs.io/en/dev/} and Gensim\footnote{https://radimrehurek.com/gensim/} to construct summaries from Gujarati text. Patel examines the preprocessing phase for text summarization of Gujarati texts, emphasizing related issues and appropriate solutions \cite{article_patel_pinkesh}.
\section{Dataset Description}
The ILSUM 2022 datasets, as provided were organized in a CSV format, with multiple columns describing each record in the file. These datasets were built using articles and headline pairs from several leading newspapers across India. The columns in the CSV files were- "id": denoting the ID for the article for unique identification, "Link": hyperlink from where the article has been extracted, "Heading": heading/title of the article, "Article": the actual content of the article, and "Summary": gold extractive summary of the article. Each article consisted, on average, of about 9 to 10 sentences, and the extractive summaries, on average, were a single sentence long.
\indent For the validation and test CSV files, there were only two columns: "id" and "Article", where it was expected to find the summary of the text present in the "Article" column. The dataset content was raw, with unnecessary punctuations and delimiters hindering the proposed pipeline, hence causing a need for efficient data cleaning. 

\indent Table \ref{Dataset details} proposes the contents of the training, validation and test datasets in terms of number of records present in each set.

\begin{table*}
\def\arraystretch{1.2}
\caption{\label{Dataset details}
Details of ILSUM 2022 datasets
}
\centering
\begin{tabular}{llll}
\hline
\textbf{} & 
\textbf{English} & 
\textbf{Hindi} &
\textbf{Gujarati} \\
\hline
Train      & 12565   & 7958  & 8731     \\ 
Validation & 898     & 569   & 606      \\ 
Test       & 4487    & 2842  & 3020     \\ 
\hline
\end{tabular}
\end{table*}

\section{Data Preparation}
The datasets pretty raw, with redundant punctuations and delimiters in the content. Hence, it was necessary to remove those so that the clean data obtained could be further tokenized and passed to the model. In addition, we remove stopwords present in the text \cite{Sarica_2021} to avoid model redundancy towards not-so-useful data and convert the text to lowercase to generalize the model perception towards the text. Out of the five columns present in the CSV file, the "id", "Link" and "Heading" columns were seemingly redundant to be taken into consideration, so we filtered out those columns for the model to get trained only on the articles and their corresponding extractive summaries. 

\indent We use the SentencePiece\footnote{https://github.com/google/sentencepiece} tokenizer for tokenizing the English, Hindi, and Gujarati article texts. SentencePiece is an unsupervised text tokenizer and detokenizer intended specifically for Neural Network-based text generation systems with a preset vocabulary size before neural model training. It extends direct training from raw sentences to incorporate subword units (e.g., byte-pair-encoding (BPE) \cite{sennrich-etal-2016-neural}) and the unigram language model. This tokenizer can be defined implicitly using Hugging Face API\footnote{https://huggingface.co/} for model fine-tuning so that both tokenization and detokenization processes can be carried out without explicit code. First, a vocabulary of all the common words in all articles is created and further utilized to quantify the text to a vectorized format. Additionally, we apply padding to the maximum sequence length in the batch so that sequences of uniform length only would be passed ahead to the model \cite{https://doi.org/10.48550/arxiv.1903.07288}.

\indent For the translation+mapping-based approach that we implement as one of the approaches for Gujarati, we first split the sentences using the full stop as a delimiter. Then, each sentence is translated to English using the Google Translate API\footnote{https://cloud.google.com/translate/}, and then the mapping is created between the original Gujarati sentence and the translated English sentence obtained. Finally, these English sentences from each paragraph are concatenated to get the final translated summaries in English.

\section{Systems implemented}

\subsection{For English}
\subsubsection{Fine-tuning PEGASUS}
PEGASUS stands for "Pre-training with Extracted Gap sentences for Abstractive Summarization", the paper for which was presented at the 2020 International Conference on Machine Learning by Zhang et al.\cite{https://doi.org/10.48550/arxiv.1912.08777}. By masking entire sentences from the text and then appending the gap sentences, the PEGASUS model yields a pseudo-summary of the input text. The PEGASUS model picks sentences that are essential to the model and removes or masks them from the input document. The model is then assigned with recovering those vital phrases, which it accomplishes by constructing the output sequence, including the critical documents entirely from the document's non-essential parts. The advantage of this technique is its self-supervision; the model may generate as many instances as there are documents without the need for human annotation, which is sometimes a bottleneck in fully supervised systems.
\\ \indent We fine-tuned the "pegasus-large" model\footnote{https://huggingface.co/google/pegasus-large} available on Hugging Face with the training dataset for English. This model is pre-trained on 350 million web pages and 1.5 billion news articles, making its accuracy state-of-the-art in text summarization research. The Hugging Face transformer library was used for fine-tuning purposes, which made the implementation easier. Since the training data was large enough, we decided to fine-tune the model for 1 epoch on the training data, along with a weight decay of 0.01, which took about 3.5 hours for the same. The inferences yielded a significant increase in ROUGE scores as observed to those obtained with the only pre-trained version of the model. 
\\ \indent To further increase the ROUGE scores, we tried experimenting with the max-tokens parameter of the model during inference generation, which is the maximum length the generated inference can have. The organizers had specified the standard value of the same to 75. We experimented with a range of max-tokens values around that range, and we got max-tokens=65 to be the ideal value for the highest ROUGE scores. 
\\ \indent We also experimented with augmenting the dataset by adding noise to each record of the dataset so that the model could predict the result better despite the noisy text present. The ROUGE score increase for the same needed to be increased, compared to the highest score we received.
 
\subsubsection{Fine-tuning BRIO}
BRIO stands for "Bringing Order to Abstractive Summarization", the paper presented in 2022 by Liu et al.\cite{liu-etal-2022-brio}. Maximum Likelihood Estimation (MLE) \cite{NEURIPS2021_d30d0f52} is often used to train summarization models. MLE presupposes that an ideal model would allocate full probability mass to the reference summary, which may result in poor performance when a model must compare numerous candidates that vary from the reference. Instead of relying on MLE training, BRIO has a contrastive learning component, enabling abstractive models to more precisely assess the likelihood of system-generated summaries.
\\ \indent We fine-tune the "Yale-LILY/brio-cnndm-uncased" version\footnote{https://huggingface.co/Yale-LILY/brio-cnndm-uncased} of the BRIO model available on Hugging Face on the English dataset. Since BRIO is an extension to the BART model, we apply BART-based tokenization to the input text, which uses SentencePiece internally. We fine-tuned the English dataset on the model for 1 epoch with a weight decay of 0.01 and even experimented further with adding noisy text to each training record. The model's performance, however, was not as good as the fine-tuned PEGASUS model mentioned earlier.

\subsubsection{Leveraging SentenceBERT for extractive summarization}
The approach was a tweaked implementation derived from the paper "Fine-tune BERT for Extractive Summarization" presented by Liu in 2019 \cite{Liu2019FinetuneBF}. Here, extractive summarization is approached as a classification problem by predicting a score between 0 and 1 for each phrase in a text, i.e., by determining whether or not it belongs to the summary. The algorithm then creates a summary based on these scores by picking the phrases with the highest scores determined by certain relevant parameters.
\\ \indent We extract sentences using the SpaCy\footnote{https://spacy.io/} library for each article in the training dataset. For every sentence in each training example, we assign a label of 1 if it belongs to the final extractive summary, else 0. The original dataset was unbalanced, as most of the sentences are unlikely to be in the summary. We augmented our dataset with new examples that balanced positive and negative examples. This annotated data, along with the labels, constitutes the input to our BERT model. \\ \indent We fine-tuned the "sentence-transformers/all-mpnet-base-v2" model\footnote{https://huggingface.co/sentence-transformers/all-mpnet-base-v2}, since it proved to be the fastest among all models available in the sentence-transformers library. We set the batch size for training to 4, and the maximum sequence length of the generated summary to 512. The learning rate for training was set to 0.00001. We fine-tuned the SentenceBERT model for 3 epochs, which took approximately 4.5 hours. The original pre-trained BERT model is modified by drop out and a dense layer on top of the BERT model to get the final output label. Finally, we get the inferences from the model by taking \textbf{two} sentences with the highest scores obtained from the BERT model, which gave an average summary length of around 70. We add only those sentences in the summary whose length is more than 25 characters.  

\subsubsection{Fine-tuning T5}
The Text-to-Text-Transfer-Transformer (T5) paradigm suggests recasting all NLP tasks as a single text-to-text format with text strings as input and output. The original text of the input and output pairs during T5 pre-training is modified by introducing noise. 
\\ \indent We fine-tuned the `mrm8488/t5-base-finetuned-summarize-news' version\footnote{https://huggingface.co/mrm8488/t5-base-finetuned-summarize-news} of the T5 model, which is pre-trained on 4515 English news articles. We fine-tuned this model on our dataset. We applied T5 tokenization to our dataset and fine-tuned the model for 20 epochs. The maximum length of the summary during the inference was set to 75.

\subsection{For Hindi}
\subsubsection{Fine-tuning IndicBART}
We used IndicBART\cite{dabre-etal-2022-indicbart}, a multilingual, sequence-to-sequence pre-trained model. The model focuses on Indic languages majorly, and English as well. IndicBART is based on the mBART architecture and provides support for 11 Indian languages, and can be used to build various natural language generation applications for tasks like machine translation and summarization. 
\\ \indent We fine-tuned the `ai4bharat/IndicBART' version\footnote{https://huggingface.co/ai4bharat/IndicBART} of IndicBART available on Hugging Face on the training dataset for Hindi. The data used for training the model was augmented by adding noise to each record of the dataset. The model gave better results better after training on such a dataset. The model was fine-tuned for 2 epochs. We also experimented with the maximum length parameter while generating the inferences. Inferences obtained with `max length' set to 60 gave the best ROUGE scores.
\subsubsection{Fine-tuning XL-Sum}
The paper titled ‘XL-Sum: Large-Scale Multilingual Abstractive Summarization for 44 Languages’ \cite{hasan-etal-2021-xl} presents a multilingual dataset as well as an mT5 checkpoint fine-tuned on the dataset and proposes fine-tuned mT5 \cite{xue-etal-2021-mt5} with XL-Sum and experimented on multilingual and low resource summarization tasks. The model was fine-tuned on the 45 languages of the XL-Sum dataset.
\\ \indent We used the 'csebuetnlp/mT5\_multilingual\_XLSum' checkpoint\footnote{https://huggingface.co/csebuetnlp/mT5\_multilingual\_XLSum} available on Hugging Face for our summarization task. To get the best results, we fine-tuned this checkpoint on the given Hindi training dataset for 2 epochs. This method gave ROUGE scores comparable to the Indic BART scores. 

\subsubsection{Fine-tuning mBART}
Pre-trained on multilingual corpora containing 25 languages, mBART (Multilingual Denoising Pre-training for Neural Machine Translation)\cite{liu2020multilingual} can be used for a wide range of tasks, including machine translation and summarization. We used the "facebook/mbart-large-cc25"\footnote{https://huggingface.co/facebook/mbart-large-cc25}, "GiordanoB/mbart-large-50-finetuned-summarization-V2"\footnote{https://huggingface.co/GiordanoB/mbart-large-50-finetuned-summarization-V2} and "ARTeLab/mbart-summarization-mlsum"\footnote{https://huggingface.co/ARTeLab/mbart-summarization-mlsum} pre-trained models on the dataset. 
\\ \indent The results obtained differed minutely. However, the "facebook/mbart-large-cc25" model gave us the best ROUGE scores; hence, we fine-tuned the model on the dataset for 1 epoch.

\subsection{For Gujarati}
\subsubsection{Translation+Mapping+PEGASUS}
We implemented the PEGASUS model for Gujarati by fine-tuning the "pegasus-large" model available on Hugging Face. As this model wasn't initially trained for the Gujarati language, we implemented translation and mapping steps to use this model for generating inferences on our Gujarati dataset.
\\ \indent First, we translated the Gujarati validation dataset to English and simultaneously stored the mapping between the English-translated sentence and the Gujarati sentence for each article in a dictionary. For translation, we used the GoogleTranslator module provided by deep-translator\footnote{https://github.com/nidhaloff/deep-translator} library. Then, we generated the inferences on the English-translated validation dataset using the PEGASUS model fine-tuned for English, the max-tokens parameter for which was set to 75 initially. Finally, the generated inferences were back-mapped to give the original Gujarati sentences. As the dataset provided was extractive, we performed the mapping and back-mapping steps mainly to keep the summaries extractive in nature. It should be noted that the translation process was only used once, and the original Gujarati text was retrieved using the mapping developed during the Gujarati to English translation process.
\\ \indent To further increase the ROUGE scores, we experimented with the max-tokens parameter of the model. We observed that the English-translated sentences were longer than the original Gujarati sentences. Therefore, we tested by increasing the max-tokens parameter, and we inferred that max-tokens set to \textbf{85} provided the highest ROUGE scores.

\subsubsection{Fine-tuning mBART}
For this approach, we used the "facebook/mbart-large-cc25"\footnote{https://huggingface.co/facebook/mbart-large-cc25} model. After applying the mBART tokenizer on the given Gujarati dataset, we fine-tuned the model for one epoch. This methodology gave us competent ROUGE scores. However, we improved our results by augmenting the dataset by adding noise to each record of the dataset to create a new record so that the model could predict better. 
\\ \indent The ROUGE scores obtained after fine-tuning the mBART model on this dataset were comparable to the Translation+Mapping+PEGASUS model.

\begin{table*}
\caption{\label{Systems ROUGE scores MS2 - English}
Results obtained in the validation set (English)
}
\centering
\begin{tabular}{llll}
\hline
\textbf{Approach Implemented} & 
\textbf{ROUGE-1} & 
\textbf{ROUGE-2} &
\textbf{ROUGE-4} \\
\hline
\textbf{Fine-tuned PEGASUS} & \textbf{0.5618} & \textbf{0.4509} & \textbf{0.4218}  \\
Fine-tuned BRIO & 0.4878 & 0.3723 & 0.3383  \\
SentenceBERT leveraged for summarization & 0.4639 & 0.3421 & 0.3156\\
Fine-tuned T5 & 0.4851 & 0.3588 & 0.3226 \\
\hline
\end{tabular}


\caption{\label{Systems ROUGE scores MS2 - Hindi}
Results obtained in the validation set (Hindi)
}
\centering
\begin{tabular}{llll}
\hline
\textbf{Approach Implemented} & 
\textbf{ROUGE-1} & 
\textbf{ROUGE-2} &
\textbf{ROUGE-4} \\
\hline
\textbf{Fine-tuned IndicBART} & \textbf{0.5536} & \textbf{0.4572} & \textbf{0.4162}  \\
Fine-tuned XL-Sum & 0.5281 & 0.4098 & 0.337  \\
Fine-tuned mBART & 0.5269 & 0.4271 & 0.3806\\
\hline
\end{tabular}


\caption{\label{Systems ROUGE scores MS2 - Gujarati}
Results obtained in the validation set (Gujarati)
}
\centering
\begin{tabular}{llll}
\hline
\textbf{Approach Implemented} & 
\textbf{ROUGE-1} & 
\textbf{ROUGE-2} &
\textbf{ROUGE-4} \\
\hline
\textbf{Translation+Mapping+PEGASUS} & \textbf{0.2028} & \textbf{0.1155} & \textbf{0.0835}  \\
Fine-tuned mBART & 0.1924 & 0.1095 & 0.0723  \\
Fine-tuned XL-Sum & 0.1718 & 0.0718 & 0.0361\\
\hline
\end{tabular}

\caption{\label{Systems ROUGE scores MS2}
The test set results for the best models
}
\centering
\begin{tabular}{lllll}
\hline
\textbf{Language} & 
\textbf{Model description} & 
\textbf{ROUGE-1} & 
\textbf{ROUGE-2} &
\textbf{ROUGE-4} \\
\hline
English & Fine-tuned PEGASUS & 0.5568 & 0.4430 & 0.4123  \\
Hindi & Fine-tuned IndicBART & 0.5559 & 0.4547 & 0.4136  \\
Gujarati & Translation+Mapping+PEGASUS & 0.2087 & 0.1192 & 0.0838\\
\hline
\end{tabular}

\end{table*}

\subsubsection{Fine-tuning XL-Sum}
\hfill\\ We used the XLSum model, an mT5 model fine-tuned on the multilingual XLSum dataset. We used the checkpoint ‘csebuetnlp/mT5\_multilingual\_XLSum’ available on Hugging Face to generate inferences on the Gujarati dataset. The model was trained for 5 epochs with the max-tokens parameter set to 75.


\section{Evaluation Metrics}
In our study, the ROUGE Score, which stands for Recall-Oriented Understudy for Gisting Assessment, was chosen as the evaluation metric \cite{lin-2004-rouge}. For our summary, we recorded ROUGE-1, ROUGE-2, and ROUGE-4 scores. ROUGE-1 calculated the unigram overlap between the candidate and reference summaries, whereas ROUGE-2 assessed the bigram similarities between the summaries. All ROUGE scores are graded out of one, with the ROUGE score closer to one, indicating more parallel with the gold summaries.

\section{Results}
Tables \ref{Systems ROUGE scores MS2 - English}, \ref{Systems ROUGE scores MS2 - Hindi} and \ref{Systems ROUGE scores MS2 - Gujarati} describe the results obtained on our approaches by testing on the validation data. Table \ref{Systems ROUGE scores MS2} describes the test data results obtained on the best validation models. We evaluated the performance of the models using ROUGE scores as the evaluation metrics. The best-performing approaches were: fine-tuned PEGASUS model with max-tokens set to 65 for English, the IndicBART model with the right-shift operation for Hindi, and the (Translation Mapping + PEGASUS) based approach for Gujarati. We achieved optimum accuracies with these approaches on both the validation and test datasets. 

\section{Conclusion and Future Work}
Thus, we have illustrated the findings of our research which we performed on the ILSUM 2022 datasets. We have experimented with text summarization on news articles written in English, Hindi, and Gujarati. We implemented pre-trained models in our research and data manipulation operations performed in some of the operations. Finally, we evaluate the ROUGE scores on the inferences obtained from each system we trained. We achieved decent accuracy on our best-performing models, with accuracies very close to SoTA accuracies. We can conclude from this analysis that there is a lot of scope for improvement in research performed for low-resource Indian languages, such as Gujarati, compared to English. The research foundation for text summarization for English is robust, as there are many pre-trained models and attention-based mechanisms that one can leverage. However, this foundation has to be scaled up drastically in the coming years for Hindi and Gujarati.
\\ \indent In the future, we plan to leverage our work on larger datasets, especially for Hindi and Gujarati, as we believe that clean and well-formatted datasets are one of the significant barriers that cause the gap between text summarization research in English and low-resource Indian languages. Furthermore, we plan to implement our approaches on high-end GPUs and use better preprocessing and tokenization techniques to shorten this research gap.

\section{Acknowledgements}
This research was accomplished as part of the L3Cube Pune mentoring program. We convey our gratitude to our L3Cube mentors for their continuous assistance and encouragement.

\bibliography{main}

\begin{thebibliography}{47}
\expandafter\ifx\csname natexlab\endcsname\relax\def\natexlab#1{#1}\fi
\providecommand{\url}[1]{\texttt{#1}}
\providecommand{\href}[2]{#2}
\providecommand{\path}[1]{#1}
\providecommand{\DOIprefix}{doi:}
\providecommand{\ArXivprefix}{arXiv:}
\providecommand{\URLprefix}{URL: }
\providecommand{\Pubmedprefix}{pmid:}
\providecommand{\doi}[1]{\href{http://dx.doi.org/#1}{\path{#1}}}
\providecommand{\Pubmed}[1]{\href{pmid:#1}{\path{#1}}}
\providecommand{\bibinfo}[2]{#2}
\ifx\xfnm\relax \def\xfnm[#1]{\unskip,\space#1}\fi
\bibitem[{Vhatkar et~al.(2020)Vhatkar, Bhattacharyya, and
  Arya}]{Vhatkar2020SurveyOT}
\bibinfo{author}{A.~Vhatkar}, \bibinfo{author}{P.~Bhattacharyya},
  \bibinfo{author}{K.~Arya},
\newblock \bibinfo{title}{Survey on text summarization},
\newblock \bibinfo{year}{2020}.
\bibitem[{Dehru et~al.(2021)Dehru, Tiwari, Aggarwal, Joshi, and
  Kartik}]{article_dehru}
\bibinfo{author}{V.~Dehru}, \bibinfo{author}{P.~Tiwari},
  \bibinfo{author}{G.~Aggarwal}, \bibinfo{author}{B.~Joshi},
  \bibinfo{author}{P.~Kartik},
\newblock \bibinfo{title}{Text summarization techniques and applications},
\newblock \bibinfo{journal}{IOP Conference Series: Materials Science and
  Engineering} \bibinfo{volume}{1099} (\bibinfo{year}{2021})
  \bibinfo{pages}{012042}. \DOIprefix\doi{10.1088/1757-899X/1099/1/012042}.
\bibitem[{Moratanch and Gopalan(2017)}]{inproceedings_extractive_survey}
\bibinfo{author}{N.~Moratanch}, \bibinfo{author}{C.~Gopalan},
\newblock \bibinfo{title}{A survey on extractive text summarization},
\newblock \bibinfo{year}{2017}, pp. \bibinfo{pages}{1--6}.
  \DOIprefix\doi{10.1109/ICCCSP.2017.7944061}.
\bibitem[{Moratanch and Gopalan(2016)}]{inproceedings_abstractive_survey}
\bibinfo{author}{N.~Moratanch}, \bibinfo{author}{C.~Gopalan},
\newblock \bibinfo{title}{A survey on abstractive text summarization},
\newblock \bibinfo{year}{2016}, pp. \bibinfo{pages}{1--7}.
  \DOIprefix\doi{10.1109/ICCPCT.2016.7530193}.
\bibitem[{Kirmani et~al.(2019)Kirmani, Hakak, mohd, and Mohd}]{inbook_hybrid}
\bibinfo{author}{M.~Kirmani}, \bibinfo{author}{N.~Hakak},
  \bibinfo{author}{M.~mohd}, \bibinfo{author}{M.~Mohd}, \bibinfo{title}{Hybrid
  Text Summarization: A Survey: Proceedings of SoCTA 2017},
  \bibinfo{year}{2019}, pp. \bibinfo{pages}{63--73}.
  \DOIprefix\doi{10.1007/978-981-13-0589-4_7}.
\bibitem[{Sahoo et~al.(2018)Sahoo, Bhoi, and Balabantaray}]{SAHOO20181228}
\bibinfo{author}{D.~Sahoo}, \bibinfo{author}{A.~Bhoi}, \bibinfo{author}{R.~C.
  Balabantaray},
\newblock \bibinfo{title}{Hybrid approach to abstractive summarization},
\newblock \bibinfo{journal}{Procedia Computer Science} \bibinfo{volume}{132}
  (\bibinfo{year}{2018}) \bibinfo{pages}{1228--1237}. \URLprefix
  \url{https://www.sciencedirect.com/science/article/pii/S1877050918307701}.
  \DOIprefix\doi{https://doi.org/10.1016/j.procs.2018.05.038},
  \bibinfo{note}{international Conference on Computational Intelligence and
  Data Science}.
\bibitem[{Luhn(1958)}]{luhn_summarization}
\bibinfo{author}{H.~P. Luhn},
\newblock \bibinfo{title}{The automatic creation of literature abstracts},
\newblock \bibinfo{journal}{IBM Journal of Research and Development}
  \bibinfo{volume}{2} (\bibinfo{year}{1958}) \bibinfo{pages}{159--165}.
  \DOIprefix\doi{10.1147/rd.22.0159}.
\bibitem[{Mohsen et~al.(2020)Mohsen, Wang, and
  Al-Sabahi}]{mohsen2020hierarchical}
\bibinfo{author}{F.~Mohsen}, \bibinfo{author}{J.~Wang},
  \bibinfo{author}{K.~Al-Sabahi},
\newblock \bibinfo{title}{A hierarchical self-attentive neural extractive
  summarizer via reinforcement learning (hsasrl)},
\newblock \bibinfo{journal}{Applied Intelligence} \bibinfo{volume}{50}
  (\bibinfo{year}{2020}) \bibinfo{pages}{2633--2646}.
\bibitem[{Xu and Durrett(2019)}]{xu2019neural}
\bibinfo{author}{J.~Xu}, \bibinfo{author}{G.~Durrett},
\newblock \bibinfo{title}{Neural extractive text summarization with syntactic
  compression},
\newblock \bibinfo{journal}{arXiv preprint arXiv:1902.00863}
  (\bibinfo{year}{2019}).
\bibitem[{Alami et~al.(2019)Alami, Meknassi, and
  En-nahnahi}]{alami2019enhancing}
\bibinfo{author}{N.~Alami}, \bibinfo{author}{M.~Meknassi},
  \bibinfo{author}{N.~En-nahnahi},
\newblock \bibinfo{title}{Enhancing unsupervised neural networks based text
  summarization with word embedding and ensemble learning},
\newblock \bibinfo{journal}{Expert systems with applications}
  \bibinfo{volume}{123} (\bibinfo{year}{2019}) \bibinfo{pages}{195--211}.
\bibitem[{Anand and Wagh(2019)}]{anand2019effective}
\bibinfo{author}{D.~Anand}, \bibinfo{author}{R.~Wagh},
\newblock \bibinfo{title}{Effective deep learning approaches for summarization
  of legal texts},
\newblock \bibinfo{journal}{Journal of King Saud University-Computer and
  Information Sciences}  (\bibinfo{year}{2019}).
\bibitem[{Han et~al.(2021)Han, Zhang, Ding, Gu, Liu, Huo, Qiu, Yao, Zhang,
  Zhang, Han, Huang, Jin, Lan, Liu, Liu, Lu, Qiu, Song, Tang, Wen, Yuan, Zhao,
  and Zhu}]{HAN2021225}
\bibinfo{author}{X.~Han}, \bibinfo{author}{Z.~Zhang},
  \bibinfo{author}{N.~Ding}, \bibinfo{author}{Y.~Gu}, \bibinfo{author}{X.~Liu},
  \bibinfo{author}{Y.~Huo}, \bibinfo{author}{J.~Qiu}, \bibinfo{author}{Y.~Yao},
  \bibinfo{author}{A.~Zhang}, \bibinfo{author}{L.~Zhang},
  \bibinfo{author}{W.~Han}, \bibinfo{author}{M.~Huang},
  \bibinfo{author}{Q.~Jin}, \bibinfo{author}{Y.~Lan}, \bibinfo{author}{Y.~Liu},
  \bibinfo{author}{Z.~Liu}, \bibinfo{author}{Z.~Lu}, \bibinfo{author}{X.~Qiu},
  \bibinfo{author}{R.~Song}, \bibinfo{author}{J.~Tang}, \bibinfo{author}{J.-R.
  Wen}, \bibinfo{author}{J.~Yuan}, \bibinfo{author}{W.~X. Zhao},
  \bibinfo{author}{J.~Zhu},
\newblock \bibinfo{title}{Pre-trained models: Past, present and future},
\newblock \bibinfo{journal}{AI Open} \bibinfo{volume}{2} (\bibinfo{year}{2021})
  \bibinfo{pages}{225--250}. \URLprefix
  \url{https://www.sciencedirect.com/science/article/pii/S2666651021000231}.
  \DOIprefix\doi{https://doi.org/10.1016/j.aiopen.2021.08.002}.
\bibitem[{Zhang et~al.(2020)Zhang, Zhao, Saleh, and
  Liu}]{10.5555/3524938.3525989}
\bibinfo{author}{J.~Zhang}, \bibinfo{author}{Y.~Zhao},
  \bibinfo{author}{M.~Saleh}, \bibinfo{author}{P.~J. Liu},
\newblock \bibinfo{title}{Pegasus: Pre-training with extracted gap-sentences
  for abstractive summarization},
\newblock in: \bibinfo{booktitle}{Proceedings of the 37th International
  Conference on Machine Learning}, ICML'20, \bibinfo{publisher}{JMLR.org},
  \bibinfo{year}{2020}.
\bibitem[{Liu et~al.(2022)Liu, Liu, Radev, and Neubig}]{liu-etal-2022-brio}
\bibinfo{author}{Y.~Liu}, \bibinfo{author}{P.~Liu}, \bibinfo{author}{D.~Radev},
  \bibinfo{author}{G.~Neubig},
\newblock \bibinfo{title}{{BRIO}: Bringing order to abstractive summarization},
\newblock in: \bibinfo{booktitle}{Proceedings of the 60th Annual Meeting of the
  Association for Computational Linguistics (Volume 1: Long Papers)},
  \bibinfo{publisher}{Association for Computational Linguistics},
  \bibinfo{address}{Dublin, Ireland}, \bibinfo{year}{2022}, pp.
  \bibinfo{pages}{2890--2903}. \URLprefix
  \url{https://aclanthology.org/2022.acl-long.207}.
  \DOIprefix\doi{10.18653/v1/2022.acl-long.207}.
\bibitem[{Raffel et~al.(2020)Raffel, Shazeer, Roberts, Lee, Narang, Matena,
  Zhou, Li, Liu et~al.}]{raffel2020exploring}
\bibinfo{author}{C.~Raffel}, \bibinfo{author}{N.~Shazeer},
  \bibinfo{author}{A.~Roberts}, \bibinfo{author}{K.~Lee},
  \bibinfo{author}{S.~Narang}, \bibinfo{author}{M.~Matena},
  \bibinfo{author}{Y.~Zhou}, \bibinfo{author}{W.~Li}, \bibinfo{author}{P.~J.
  Liu}, et~al.,
\newblock \bibinfo{title}{Exploring the limits of transfer learning with a
  unified text-to-text transformer.},
\newblock \bibinfo{journal}{J. Mach. Learn. Res.} \bibinfo{volume}{21}
  (\bibinfo{year}{2020}) \bibinfo{pages}{1--67}.
\bibitem[{Reimers and Gurevych(2019)}]{reimers2019sentence}
\bibinfo{author}{N.~Reimers}, \bibinfo{author}{I.~Gurevych},
\newblock \bibinfo{title}{Sentence-bert: Sentence embeddings using siamese
  bert-networks},
\newblock \bibinfo{journal}{arXiv preprint arXiv:1908.10084}
  (\bibinfo{year}{2019}).
\bibitem[{Liu(2019)}]{liu2019fine}
\bibinfo{author}{Y.~Liu},
\newblock \bibinfo{title}{Fine-tune bert for extractive summarization},
\newblock \bibinfo{journal}{arXiv preprint arXiv:1903.10318}
  (\bibinfo{year}{2019}).
\bibitem[{Dabre et~al.(2022)Dabre, Shrotriya, Kunchukuttan, Puduppully, Khapra,
  and Kumar}]{dabre-etal-2022-indicbart}
\bibinfo{author}{R.~Dabre}, \bibinfo{author}{H.~Shrotriya},
  \bibinfo{author}{A.~Kunchukuttan}, \bibinfo{author}{R.~Puduppully},
  \bibinfo{author}{M.~Khapra}, \bibinfo{author}{P.~Kumar},
\newblock \bibinfo{title}{{I}ndic{BART}: A pre-trained model for indic natural
  language generation},
\newblock in: \bibinfo{booktitle}{Findings of the Association for Computational
  Linguistics: ACL 2022}, \bibinfo{publisher}{Association for Computational
  Linguistics}, \bibinfo{address}{Dublin, Ireland}, \bibinfo{year}{2022}, pp.
  \bibinfo{pages}{1849--1863}. \URLprefix
  \url{https://aclanthology.org/2022.findings-acl.145}.
  \DOIprefix\doi{10.18653/v1/2022.findings-acl.145}.
\bibitem[{Hasan et~al.(2021)Hasan, Bhattacharjee, Islam, Mubasshir, Li, Kang,
  Rahman, and Shahriyar}]{hasan-etal-2021-xl}
\bibinfo{author}{T.~Hasan}, \bibinfo{author}{A.~Bhattacharjee},
  \bibinfo{author}{M.~S. Islam}, \bibinfo{author}{K.~Mubasshir},
  \bibinfo{author}{Y.-F. Li}, \bibinfo{author}{Y.-B. Kang},
  \bibinfo{author}{M.~S. Rahman}, \bibinfo{author}{R.~Shahriyar},
\newblock \bibinfo{title}{{XL}-sum: Large-scale multilingual abstractive
  summarization for 44 languages},
\newblock in: \bibinfo{booktitle}{Findings of the Association for Computational
  Linguistics: ACL-IJCNLP 2021}, \bibinfo{publisher}{Association for
  Computational Linguistics}, \bibinfo{address}{Online}, \bibinfo{year}{2021},
  pp. \bibinfo{pages}{4693--4703}. \URLprefix
  \url{https://aclanthology.org/2021.findings-acl.413}.
  \DOIprefix\doi{10.18653/v1/2021.findings-acl.413}.
\bibitem[{Liu et~al.(2020)Liu, Gu, Goyal, Li, Edunov, Ghazvininejad, Lewis, and
  Zettlemoyer}]{liu2020multilingual}
\bibinfo{author}{Y.~Liu}, \bibinfo{author}{J.~Gu}, \bibinfo{author}{N.~Goyal},
  \bibinfo{author}{X.~Li}, \bibinfo{author}{S.~Edunov},
  \bibinfo{author}{M.~Ghazvininejad}, \bibinfo{author}{M.~Lewis},
  \bibinfo{author}{L.~Zettlemoyer},
\newblock \bibinfo{title}{Multilingual denoising pre-training for neural
  machine translation},
\newblock \bibinfo{journal}{Transactions of the Association for Computational
  Linguistics} \bibinfo{volume}{8} (\bibinfo{year}{2020})
  \bibinfo{pages}{726--742}.
\bibitem[{Baxendale(1958)}]{5392648}
\bibinfo{author}{P.~B. Baxendale},
\newblock \bibinfo{title}{Machine-made index for technical literature—an
  experiment},
\newblock \bibinfo{journal}{IBM Journal of Research and Development}
  \bibinfo{volume}{2} (\bibinfo{year}{1958}) \bibinfo{pages}{354--361}.
  \DOIprefix\doi{10.1147/rd.24.0354}.
\bibitem[{Oliveira et~al.(2016)Oliveira, Ferreira, Lima, Lins, Freitas, Riss,
  and Simske}]{10.1016/j.eswa.2016.08.030}
\bibinfo{author}{H.~Oliveira}, \bibinfo{author}{R.~Ferreira},
  \bibinfo{author}{R.~Lima}, \bibinfo{author}{R.~D. Lins},
  \bibinfo{author}{F.~Freitas}, \bibinfo{author}{M.~Riss},
  \bibinfo{author}{S.~J. Simske},
\newblock \bibinfo{title}{Assessing shallow sentence scoring techniques and
  combinations for single and multi-document summarization},
\newblock \bibinfo{journal}{Expert Syst. Appl.} \bibinfo{volume}{65}
  (\bibinfo{year}{2016}) \bibinfo{pages}{68–86}. \URLprefix
  \url{https://doi.org/10.1016/j.eswa.2016.08.030}.
  \DOIprefix\doi{10.1016/j.eswa.2016.08.030}.
\bibitem[{Aitken et~al.(2021)Aitken, Ramasesh, Cao, and
  Maheswaranathan}]{https://doi.org/10.48550/arxiv.2110.15253}
\bibinfo{author}{K.~Aitken}, \bibinfo{author}{V.~V. Ramasesh},
  \bibinfo{author}{Y.~Cao}, \bibinfo{author}{N.~Maheswaranathan},
  \bibinfo{title}{Understanding how encoder-decoder architectures attend},
  \bibinfo{year}{2021}. \URLprefix \url{https://arxiv.org/abs/2110.15253}.
  \DOIprefix\doi{10.48550/ARXIV.2110.15253}.
\bibitem[{Vaswani et~al.(2017)Vaswani, Shazeer, Parmar, Uszkoreit, Jones,
  Gomez, Kaiser, and Polosukhin}]{NIPS2017_3f5ee243}
\bibinfo{author}{A.~Vaswani}, \bibinfo{author}{N.~Shazeer},
  \bibinfo{author}{N.~Parmar}, \bibinfo{author}{J.~Uszkoreit},
  \bibinfo{author}{L.~Jones}, \bibinfo{author}{A.~N. Gomez},
  \bibinfo{author}{L.~u. Kaiser}, \bibinfo{author}{I.~Polosukhin},
\newblock \bibinfo{title}{Attention is all you need},
\newblock in: \bibinfo{editor}{I.~Guyon}, \bibinfo{editor}{U.~V. Luxburg},
  \bibinfo{editor}{S.~Bengio}, \bibinfo{editor}{H.~Wallach},
  \bibinfo{editor}{R.~Fergus}, \bibinfo{editor}{S.~Vishwanathan},
  \bibinfo{editor}{R.~Garnett} (Eds.), \bibinfo{booktitle}{Advances in Neural
  Information Processing Systems}, volume~\bibinfo{volume}{30},
  \bibinfo{publisher}{Curran Associates, Inc.}, \bibinfo{year}{2017}.
  \URLprefix
  \url{https://proceedings.neurips.cc/paper/2017/file/3f5ee243547dee91fbd053c1c4a845aa-Paper.pdf}.
\bibitem[{Yu(2017)}]{Yu2017SummarizationWA}
\bibinfo{author}{H.~Yu},
\newblock \bibinfo{title}{Summarization with attention-based deep recurrent
  neural networks},
\newblock \bibinfo{year}{2017}.
\bibitem[{Hochreiter and Schmidhuber(1997)}]{LSTM_1}
\bibinfo{author}{S.~Hochreiter}, \bibinfo{author}{J.~Schmidhuber},
\newblock \bibinfo{title}{Long short-term memory},
\newblock \bibinfo{journal}{Neural computation} \bibinfo{volume}{9}
  (\bibinfo{year}{1997}) \bibinfo{pages}{1735--80}.
  \DOIprefix\doi{10.1162/neco.1997.9.8.1735}.
\bibitem[{Chung et~al.(2014)Chung, Gulcehre, Cho, and
  Bengio}]{https://doi.org/10.48550/arxiv.1412.3555}
\bibinfo{author}{J.~Chung}, \bibinfo{author}{C.~Gulcehre},
  \bibinfo{author}{K.~Cho}, \bibinfo{author}{Y.~Bengio},
  \bibinfo{title}{Empirical evaluation of gated recurrent neural networks on
  sequence modeling}, \bibinfo{year}{2014}. \URLprefix
  \url{https://arxiv.org/abs/1412.3555}.
  \DOIprefix\doi{10.48550/ARXIV.1412.3555}.
\bibitem[{Lewis et~al.(2020)Lewis, Liu, Goyal, Ghazvininejad, Mohamed, Levy,
  Stoyanov, and Zettlemoyer}]{lewis-etal-2020-bart}
\bibinfo{author}{M.~Lewis}, \bibinfo{author}{Y.~Liu},
  \bibinfo{author}{N.~Goyal}, \bibinfo{author}{M.~Ghazvininejad},
  \bibinfo{author}{A.~Mohamed}, \bibinfo{author}{O.~Levy},
  \bibinfo{author}{V.~Stoyanov}, \bibinfo{author}{L.~Zettlemoyer},
\newblock \bibinfo{title}{{BART}: Denoising sequence-to-sequence pre-training
  for natural language generation, translation, and comprehension},
\newblock in: \bibinfo{booktitle}{Proceedings of the 58th Annual Meeting of the
  Association for Computational Linguistics}, \bibinfo{publisher}{Association
  for Computational Linguistics}, \bibinfo{address}{Online},
  \bibinfo{year}{2020}, pp. \bibinfo{pages}{7871--7880}. \URLprefix
  \url{https://aclanthology.org/2020.acl-main.703}.
  \DOIprefix\doi{10.18653/v1/2020.acl-main.703}.
\bibitem[{Lebanoff et~al.(2020)Lebanoff, Dernoncourt, Kim, Chang, and
  Liu}]{lebanoff-etal-2020-cascade}
\bibinfo{author}{L.~Lebanoff}, \bibinfo{author}{F.~Dernoncourt},
  \bibinfo{author}{D.~S. Kim}, \bibinfo{author}{W.~Chang},
  \bibinfo{author}{F.~Liu},
\newblock \bibinfo{title}{A cascade approach to neural abstractive
  summarization with content selection and fusion},
\newblock in: \bibinfo{booktitle}{Proceedings of the 1st Conference of the
  Asia-Pacific Chapter of the Association for Computational Linguistics and the
  10th International Joint Conference on Natural Language Processing},
  \bibinfo{publisher}{Association for Computational Linguistics},
  \bibinfo{address}{Suzhou, China}, \bibinfo{year}{2020}, pp.
  \bibinfo{pages}{529--535}. \URLprefix
  \url{https://aclanthology.org/2020.aacl-main.52}.
\bibitem[{Brown et~al.(2020)Brown, Mann, Ryder, Subbiah, Kaplan, Dhariwal,
  Neelakantan, Shyam, Sastry, Askell, Agarwal, Herbert-Voss, Krueger, Henighan,
  Child, Ramesh, Ziegler, Wu, Winter, Hesse, Chen, Sigler, Litwin, Gray, Chess,
  Clark, Berner, McCandlish, Radford, Sutskever, and
  Amodei}]{NEURIPS2020_1457c0d6}
\bibinfo{author}{T.~Brown}, \bibinfo{author}{B.~Mann},
  \bibinfo{author}{N.~Ryder}, \bibinfo{author}{M.~Subbiah},
  \bibinfo{author}{J.~D. Kaplan}, \bibinfo{author}{P.~Dhariwal},
  \bibinfo{author}{A.~Neelakantan}, \bibinfo{author}{P.~Shyam},
  \bibinfo{author}{G.~Sastry}, \bibinfo{author}{A.~Askell},
  \bibinfo{author}{S.~Agarwal}, \bibinfo{author}{A.~Herbert-Voss},
  \bibinfo{author}{G.~Krueger}, \bibinfo{author}{T.~Henighan},
  \bibinfo{author}{R.~Child}, \bibinfo{author}{A.~Ramesh},
  \bibinfo{author}{D.~Ziegler}, \bibinfo{author}{J.~Wu},
  \bibinfo{author}{C.~Winter}, \bibinfo{author}{C.~Hesse},
  \bibinfo{author}{M.~Chen}, \bibinfo{author}{E.~Sigler},
  \bibinfo{author}{M.~Litwin}, \bibinfo{author}{S.~Gray},
  \bibinfo{author}{B.~Chess}, \bibinfo{author}{J.~Clark},
  \bibinfo{author}{C.~Berner}, \bibinfo{author}{S.~McCandlish},
  \bibinfo{author}{A.~Radford}, \bibinfo{author}{I.~Sutskever},
  \bibinfo{author}{D.~Amodei},
\newblock \bibinfo{title}{Language models are few-shot learners},
\newblock in: \bibinfo{editor}{H.~Larochelle}, \bibinfo{editor}{M.~Ranzato},
  \bibinfo{editor}{R.~Hadsell}, \bibinfo{editor}{M.~Balcan},
  \bibinfo{editor}{H.~Lin} (Eds.), \bibinfo{booktitle}{Advances in Neural
  Information Processing Systems}, volume~\bibinfo{volume}{33},
  \bibinfo{publisher}{Curran Associates, Inc.}, \bibinfo{year}{2020}, pp.
  \bibinfo{pages}{1877--1901}. \URLprefix
  \url{https://proceedings.neurips.cc/paper/2020/file/1457c0d6bfcb4967418bfb8ac142f64a-Paper.pdf}.
\bibitem[{Chowdhery et~al.(2022)Chowdhery, Narang, Devlin, Bosma, Mishra,
  Roberts, Barham, Chung, Sutton, Gehrmann, Schuh, Shi, Tsvyashchenko, Maynez,
  Rao, Barnes, Tay, Shazeer, Prabhakaran, Reif, Du, Hutchinson, Pope, Bradbury,
  Austin, Isard, Gur-Ari, Yin, Duke, Levskaya, Ghemawat, Dev, Michalewski,
  Garcia, Misra, Robinson, Fedus, Zhou, Ippolito, Luan, Lim, Zoph, Spiridonov,
  Sepassi, Dohan, Agrawal, Omernick, Dai, Pillai, Pellat, Lewkowycz, Moreira,
  Child, Polozov, Lee, Zhou, Wang, Saeta, Diaz, Firat, Catasta, Wei,
  Meier-Hellstern, Eck, Dean, Petrov, and
  Fiedel}]{https://doi.org/10.48550/arxiv.2204.02311}
\bibinfo{author}{A.~Chowdhery}, \bibinfo{author}{S.~Narang},
  \bibinfo{author}{J.~Devlin}, \bibinfo{author}{M.~Bosma},
  \bibinfo{author}{G.~Mishra}, \bibinfo{author}{A.~Roberts},
  \bibinfo{author}{P.~Barham}, \bibinfo{author}{H.~W. Chung},
  \bibinfo{author}{C.~Sutton}, \bibinfo{author}{S.~Gehrmann},
  \bibinfo{author}{P.~Schuh}, \bibinfo{author}{K.~Shi},
  \bibinfo{author}{S.~Tsvyashchenko}, \bibinfo{author}{J.~Maynez},
  \bibinfo{author}{A.~Rao}, \bibinfo{author}{P.~Barnes},
  \bibinfo{author}{Y.~Tay}, \bibinfo{author}{N.~Shazeer},
  \bibinfo{author}{V.~Prabhakaran}, \bibinfo{author}{E.~Reif},
  \bibinfo{author}{N.~Du}, \bibinfo{author}{B.~Hutchinson},
  \bibinfo{author}{R.~Pope}, \bibinfo{author}{J.~Bradbury},
  \bibinfo{author}{J.~Austin}, \bibinfo{author}{M.~Isard},
  \bibinfo{author}{G.~Gur-Ari}, \bibinfo{author}{P.~Yin},
  \bibinfo{author}{T.~Duke}, \bibinfo{author}{A.~Levskaya},
  \bibinfo{author}{S.~Ghemawat}, \bibinfo{author}{S.~Dev},
  \bibinfo{author}{H.~Michalewski}, \bibinfo{author}{X.~Garcia},
  \bibinfo{author}{V.~Misra}, \bibinfo{author}{K.~Robinson},
  \bibinfo{author}{L.~Fedus}, \bibinfo{author}{D.~Zhou},
  \bibinfo{author}{D.~Ippolito}, \bibinfo{author}{D.~Luan},
  \bibinfo{author}{H.~Lim}, \bibinfo{author}{B.~Zoph},
  \bibinfo{author}{A.~Spiridonov}, \bibinfo{author}{R.~Sepassi},
  \bibinfo{author}{D.~Dohan}, \bibinfo{author}{S.~Agrawal},
  \bibinfo{author}{M.~Omernick}, \bibinfo{author}{A.~M. Dai},
  \bibinfo{author}{T.~S. Pillai}, \bibinfo{author}{M.~Pellat},
  \bibinfo{author}{A.~Lewkowycz}, \bibinfo{author}{E.~Moreira},
  \bibinfo{author}{R.~Child}, \bibinfo{author}{O.~Polozov},
  \bibinfo{author}{K.~Lee}, \bibinfo{author}{Z.~Zhou},
  \bibinfo{author}{X.~Wang}, \bibinfo{author}{B.~Saeta},
  \bibinfo{author}{M.~Diaz}, \bibinfo{author}{O.~Firat},
  \bibinfo{author}{M.~Catasta}, \bibinfo{author}{J.~Wei},
  \bibinfo{author}{K.~Meier-Hellstern}, \bibinfo{author}{D.~Eck},
  \bibinfo{author}{J.~Dean}, \bibinfo{author}{S.~Petrov},
  \bibinfo{author}{N.~Fiedel}, \bibinfo{title}{Palm: Scaling language modeling
  with pathways}, \bibinfo{year}{2022}. \URLprefix
  \url{https://arxiv.org/abs/2204.02311}.
  \DOIprefix\doi{10.48550/ARXIV.2204.02311}.
\bibitem[{Sanh et~al.(2022)Sanh, Webson, Raffel, Bach, Sutawika, Alyafeai,
  Chaffin, Stiegler, Raja, Dey, Bari, Xu, Thakker, Sharma, Szczechla, Kim,
  Chhablani, Nayak, Datta, Chang, Jiang, Wang, Manica, Shen, Yong, Pandey,
  Bawden, Wang, Neeraj, Rozen, Sharma, Santilli, Fevry, Fries, Teehan, Scao,
  Biderman, Gao, Wolf, and Rush}]{sanh2022multitask}
\bibinfo{author}{V.~Sanh}, \bibinfo{author}{A.~Webson},
  \bibinfo{author}{C.~Raffel}, \bibinfo{author}{S.~Bach},
  \bibinfo{author}{L.~Sutawika}, \bibinfo{author}{Z.~Alyafeai},
  \bibinfo{author}{A.~Chaffin}, \bibinfo{author}{A.~Stiegler},
  \bibinfo{author}{A.~Raja}, \bibinfo{author}{M.~Dey}, \bibinfo{author}{M.~S.
  Bari}, \bibinfo{author}{C.~Xu}, \bibinfo{author}{U.~Thakker},
  \bibinfo{author}{S.~S. Sharma}, \bibinfo{author}{E.~Szczechla},
  \bibinfo{author}{T.~Kim}, \bibinfo{author}{G.~Chhablani},
  \bibinfo{author}{N.~Nayak}, \bibinfo{author}{D.~Datta},
  \bibinfo{author}{J.~Chang}, \bibinfo{author}{M.~T.-J. Jiang},
  \bibinfo{author}{H.~Wang}, \bibinfo{author}{M.~Manica},
  \bibinfo{author}{S.~Shen}, \bibinfo{author}{Z.~X. Yong},
  \bibinfo{author}{H.~Pandey}, \bibinfo{author}{R.~Bawden},
  \bibinfo{author}{T.~Wang}, \bibinfo{author}{T.~Neeraj},
  \bibinfo{author}{J.~Rozen}, \bibinfo{author}{A.~Sharma},
  \bibinfo{author}{A.~Santilli}, \bibinfo{author}{T.~Fevry},
  \bibinfo{author}{J.~A. Fries}, \bibinfo{author}{R.~Teehan},
  \bibinfo{author}{T.~L. Scao}, \bibinfo{author}{S.~Biderman},
  \bibinfo{author}{L.~Gao}, \bibinfo{author}{T.~Wolf}, \bibinfo{author}{A.~M.
  Rush},
\newblock \bibinfo{title}{Multitask prompted training enables zero-shot task
  generalization},
\newblock in: \bibinfo{booktitle}{International Conference on Learning
  Representations}, \bibinfo{year}{2022}. \URLprefix
  \url{https://openreview.net/forum?id=9Vrb9D0WI4}.
\bibitem[{Wang et~al.(2020)Wang, Liu, Zheng, Qiu, and
  Huang}]{wang-etal-2020-heterogeneous}
\bibinfo{author}{D.~Wang}, \bibinfo{author}{P.~Liu},
  \bibinfo{author}{Y.~Zheng}, \bibinfo{author}{X.~Qiu},
  \bibinfo{author}{X.~Huang},
\newblock \bibinfo{title}{Heterogeneous graph neural networks for extractive
  document summarization},
\newblock in: \bibinfo{booktitle}{Proceedings of the 58th Annual Meeting of the
  Association for Computational Linguistics}, \bibinfo{publisher}{Association
  for Computational Linguistics}, \bibinfo{address}{Online},
  \bibinfo{year}{2020}, pp. \bibinfo{pages}{6209--6219}. \URLprefix
  \url{https://aclanthology.org/2020.acl-main.553}.
  \DOIprefix\doi{10.18653/v1/2020.acl-main.553}.
\bibitem[{Zhong et~al.(2020)Zhong, Liu, Chen, Wang, Qiu, and
  Huang}]{zhong-etal-2020-extractive}
\bibinfo{author}{M.~Zhong}, \bibinfo{author}{P.~Liu},
  \bibinfo{author}{Y.~Chen}, \bibinfo{author}{D.~Wang},
  \bibinfo{author}{X.~Qiu}, \bibinfo{author}{X.~Huang},
\newblock \bibinfo{title}{Extractive summarization as text matching},
\newblock in: \bibinfo{booktitle}{Proceedings of the 58th Annual Meeting of the
  Association for Computational Linguistics}, \bibinfo{publisher}{Association
  for Computational Linguistics}, \bibinfo{address}{Online},
  \bibinfo{year}{2020}, pp. \bibinfo{pages}{6197--6208}. \URLprefix
  \url{https://aclanthology.org/2020.acl-main.552}.
  \DOIprefix\doi{10.18653/v1/2020.acl-main.552}.
\bibitem[{Kumar et~al.(2015)Kumar, Yadav, and
  Sharma}]{10.1007/978-81-322-2250-7_29}
\bibinfo{author}{K.~V. Kumar}, \bibinfo{author}{D.~Yadav},
  \bibinfo{author}{A.~Sharma},
\newblock \bibinfo{title}{Graph based technique for hindi text summarization},
\newblock in: \bibinfo{editor}{J.~K. Mandal}, \bibinfo{editor}{S.~C.
  Satapathy}, \bibinfo{editor}{M.~Kumar~Sanyal}, \bibinfo{editor}{P.~P.
  Sarkar}, \bibinfo{editor}{A.~Mukhopadhyay} (Eds.),
  \bibinfo{booktitle}{Information Systems Design and Intelligent Applications},
  \bibinfo{publisher}{Springer India}, \bibinfo{address}{New Delhi},
  \bibinfo{year}{2015}, pp. \bibinfo{pages}{301--310}.
\bibitem[{Gulati and Sawarkar(2017)}]{Gulati2017ANT}
\bibinfo{author}{A.~N. Gulati}, \bibinfo{author}{S.~D. Sawarkar},
\newblock \bibinfo{title}{A novel technique for multidocument hindi text
  summarization},
\newblock \bibinfo{journal}{2017 International Conference on Nascent
  Technologies in Engineering (ICNTE)}  (\bibinfo{year}{2017})
  \bibinfo{pages}{1--6}.
\bibitem[{Gupta and Garg(2016)}]{Gupta2016TextSO}
\bibinfo{author}{M.~Gupta}, \bibinfo{author}{N.~K. Garg},
\newblock \bibinfo{title}{Text summarization of hindi documents using rule
  based approach},
\newblock \bibinfo{journal}{2016 International Conference on Micro-Electronics
  and Telecommunication Engineering (ICMETE)}  (\bibinfo{year}{2016})
  \bibinfo{pages}{366--370}.
\bibitem[{Jain et~al.(2022)Jain, Arora, Morato, Yadav, Kumar~K, Automatic,
  Szymanski, Mora, Logofătu, Sobecki, and Jain}]{article_jain}
\bibinfo{author}{A.~Jain}, \bibinfo{author}{A.~Arora},
  \bibinfo{author}{J.~Morato}, \bibinfo{author}{D.~Yadav},
  \bibinfo{author}{V.~Kumar~K}, \bibinfo{author}{Automatic},
  \bibinfo{author}{J.~Szymanski}, \bibinfo{author}{H.~Mora},
  \bibinfo{author}{D.~Logofătu}, \bibinfo{author}{A.~Sobecki},
  \bibinfo{author}{D.~Jain},
\newblock \bibinfo{title}{Automatic text summarization for hindi using real
  coded genetic algorithm},
\newblock \bibinfo{journal}{Applied Sciences} \bibinfo{volume}{12}
  (\bibinfo{year}{2022}). \DOIprefix\doi{10.3390/app12136584}.
\bibitem[{Patel(2014)}]{article_patel_pinkesh}
\bibinfo{author}{P.~Patel},
\newblock \bibinfo{title}{Pre-processing phase of text summarization based on
  gujarati language},
\newblock \bibinfo{journal}{International Journal of Innovative Research in
  Computer Science and Technology} \bibinfo{volume}{ISSN}
  (\bibinfo{year}{2014}) \bibinfo{pages}{2347--5552}.
\bibitem[{Sarica and Luo(2021)}]{Sarica_2021}
\bibinfo{author}{S.~Sarica}, \bibinfo{author}{J.~Luo},
\newblock \bibinfo{title}{Stopwords in technical language processing},
\newblock \bibinfo{journal}{{PLOS} {ONE}} \bibinfo{volume}{16}
  (\bibinfo{year}{2021}) \bibinfo{pages}{e0254937}. \URLprefix
  \url{https://doi.org/10.1371%2Fjournal.pone.0254937}.
  \DOIprefix\doi{10.1371/journal.pone.0254937}.
\bibitem[{Sennrich et~al.(2016)Sennrich, Haddow, and
  Birch}]{sennrich-etal-2016-neural}
\bibinfo{author}{R.~Sennrich}, \bibinfo{author}{B.~Haddow},
  \bibinfo{author}{A.~Birch},
\newblock \bibinfo{title}{Neural machine translation of rare words with subword
  units},
\newblock in: \bibinfo{booktitle}{Proceedings of the 54th Annual Meeting of the
  Association for Computational Linguistics (Volume 1: Long Papers)},
  \bibinfo{publisher}{Association for Computational Linguistics},
  \bibinfo{address}{Berlin, Germany}, \bibinfo{year}{2016}, pp.
  \bibinfo{pages}{1715--1725}. \URLprefix
  \url{https://aclanthology.org/P16-1162}.
  \DOIprefix\doi{10.18653/v1/P16-1162}.
\bibitem[{Dwarampudi and
  Reddy(2019)}]{https://doi.org/10.48550/arxiv.1903.07288}
\bibinfo{author}{M.~Dwarampudi}, \bibinfo{author}{N.~V.~S. Reddy},
  \bibinfo{title}{Effects of padding on lstms and cnns}, \bibinfo{year}{2019}.
  \URLprefix \url{https://arxiv.org/abs/1903.07288}.
  \DOIprefix\doi{10.48550/ARXIV.1903.07288}.
\bibitem[{Zhang et~al.(2019)Zhang, Zhao, Saleh, and
  Liu}]{https://doi.org/10.48550/arxiv.1912.08777}
\bibinfo{author}{J.~Zhang}, \bibinfo{author}{Y.~Zhao},
  \bibinfo{author}{M.~Saleh}, \bibinfo{author}{P.~J. Liu},
  \bibinfo{title}{Pegasus: Pre-training with extracted gap-sentences for
  abstractive summarization}, \bibinfo{year}{2019}. \URLprefix
  \url{https://arxiv.org/abs/1912.08777}.
  \DOIprefix\doi{10.48550/ARXIV.1912.08777}.
\bibitem[{Wang et~al.(2021)Wang, Chen, Saxon, and Wang}]{NEURIPS2021_d30d0f52}
\bibinfo{author}{X.~Wang}, \bibinfo{author}{W.~Chen},
  \bibinfo{author}{M.~Saxon}, \bibinfo{author}{W.~Y. Wang},
\newblock \bibinfo{title}{Counterfactual maximum likelihood estimation for
  training deep networks},
\newblock in: \bibinfo{editor}{M.~Ranzato}, \bibinfo{editor}{A.~Beygelzimer},
  \bibinfo{editor}{Y.~Dauphin}, \bibinfo{editor}{P.~Liang},
  \bibinfo{editor}{J.~W. Vaughan} (Eds.), \bibinfo{booktitle}{Advances in
  Neural Information Processing Systems}, volume~\bibinfo{volume}{34},
  \bibinfo{publisher}{Curran Associates, Inc.}, \bibinfo{year}{2021}, pp.
  \bibinfo{pages}{25072--25085}. \URLprefix
  \url{https://proceedings.neurips.cc/paper/2021/file/d30d0f522a86b3665d8e3a9a91472e28-Paper.pdf}.
\bibitem[{Liu(2019)}]{Liu2019FinetuneBF}
\bibinfo{author}{Y.~Liu},
\newblock \bibinfo{title}{Fine-tune bert for extractive summarization},
\newblock \bibinfo{journal}{ArXiv} \bibinfo{volume}{abs/1903.10318}
  (\bibinfo{year}{2019}).
\bibitem[{Xue et~al.(2021)Xue, Constant, Roberts, Kale, Al-Rfou, Siddhant,
  Barua, and Raffel}]{xue-etal-2021-mt5}
\bibinfo{author}{L.~Xue}, \bibinfo{author}{N.~Constant},
  \bibinfo{author}{A.~Roberts}, \bibinfo{author}{M.~Kale},
  \bibinfo{author}{R.~Al-Rfou}, \bibinfo{author}{A.~Siddhant},
  \bibinfo{author}{A.~Barua}, \bibinfo{author}{C.~Raffel},
\newblock \bibinfo{title}{m{T}5: A massively multilingual pre-trained
  text-to-text transformer},
\newblock in: \bibinfo{booktitle}{Proceedings of the 2021 Conference of the
  North American Chapter of the Association for Computational Linguistics:
  Human Language Technologies}, \bibinfo{publisher}{Association for
  Computational Linguistics}, \bibinfo{address}{Online}, \bibinfo{year}{2021},
  pp. \bibinfo{pages}{483--498}. \URLprefix
  \url{https://aclanthology.org/2021.naacl-main.41}.
  \DOIprefix\doi{10.18653/v1/2021.naacl-main.41}.
\bibitem[{Lin(2004)}]{lin-2004-rouge}
\bibinfo{author}{C.-Y. Lin},
\newblock \bibinfo{title}{{ROUGE}: A package for automatic evaluation of
  summaries},
\newblock in: \bibinfo{booktitle}{Text Summarization Branches Out},
  \bibinfo{publisher}{Association for Computational Linguistics},
  \bibinfo{address}{Barcelona, Spain}, \bibinfo{year}{2004}, pp.
  \bibinfo{pages}{74--81}. \URLprefix \url{https://aclanthology.org/W04-1013}.

\end{thebibliography}

\end{document}